# A PROTOTYPE FOR EDUCATIONAL PLANNING USING COURSE CONSTRAINTS TO SIMULATE STUDENT POPULATIONS

THANASIS HADZILACOS

*Open University of Cyprus, PO box 24801,
1304, Nicosia, Cyprus
thh @ouc.ac.cy*

DIMITRIS KALLES

*Hellenic Open University, Tsamadou 13-15,
Patras, 26222, Greece
kalles@eap.gr*

DIMITRIS KOUMANAKOS

VASSILIS MITSIONIS

Distance learning universities usually afford their students the flexibility to advance their studies at their own pace. This can lead to a considerable fluctuation of student populations within a program's courses, possibly affecting the academic viability of a program as well as the related required resources. Providing a method that estimates this population could be of substantial help to university management and academic personnel. We describe how to use course precedence constraints to calculate alternative tuition paths and then use Markov models to estimate future populations. In doing so, we identify key issues of a large scale potential deployment.

## 1. Introduction

Distance learning universities usually afford their students the flexibility to advance their studies at their own pace. This can lead to considerable fluctuation of student populations within a program's courses, possibly affecting the academic viability of a program as well as the resources that have to be budgeted and administered. Providing a method that could guide management and academic personnel towards estimating this population could be of substantial administrative value.[1,2]

Such fluctuations also occur in the Hellenic Open University where students' personal circumstances may easily change within short periods, mostly due to family and employment reasons. Moreover, as in most similar universities, significant drop-out rates are recorded in some programs, usually as a result of failure in a junior year. While understanding and addressing the reasons of failure is an educational problem, drop-out also amplifies the administrative consequences of unexpected fluctuations in the student population.

In HOU, administrative aspects that are mostly affected by the student population include tutor contract renewal, tutoring venue rental and allocation, the procurement and distribution of educational material, and the development and operation of (mostly IT) infrastructure. Cost consciousness is essential during planning for several-year contracts.

In this paper we present a method to estimate student populations based on course precedence constraints. These constraints are used to calculate alternative tuition paths for students, based on data about past enrolments and exam successes. The initial motivation was to estimate the number of students in the program, so as to better argue about how many students may be admitted at registration.

The rest of the paper is structured in three sections. We next present the specification and implementation of the simulation in what constitutes the core of the paper. Following that, we identify the issues that we need to resolve before we field our approach at a larger scale. Finally, while concluding, we also briefly reflect on the political aspects of using simulation for educational planning.

## 2. Specification and Implementation of the Simulation

In our simulation, we focus on a Master's conversion program in Information Systems, featuring five taught modules, four of which must be completed to proceed to a thesis. Of those modules, one is a compulsory and demanding introduction to the program with recorded success rates of about 50-70% and drop-out being the usual path after failure.

A module is the basic educational unit at HOU. It runs for about ten months and is the equivalent of about 3-4 conventional university semester courses. A typical class contains about ten to thirty students (depending on geographical distribution) and is assigned to a tutor. All tutors of classes of the same module collaborate on various module aspects. Each student must turn in some written assignments (typically six), which contribute towards the final grade, before sitting a written exam. Students may not sit the written exam if they do not achieve a pass grade in the assignments they turn in.

For ease of reference, we will use the short module codes in the rest of this paper: 50, 51, 60, 61 and 62. Modules 50 and 51 are junior modules and are compulsory. Modules 60, 61 and 62 are senior modules and any two of them may be selected. A student may attend at most two modules per year; when four modules are successfully completed, the student may proceed to a final year thesis. The basic precedence constraints are:

- Module 50 must be successfully cleared before enrolling in module 60 or 61,
- Module 50 must be the first to be selected,
- Module 51 must be successfully cleared before enrolling in module 62,
- Enrolment in any senior module cannot be prior to enrolment in any junior module.

Moving from registration to graduation can be cast as a search problem, where any legal path from a start state (registration) to an end state (graduation or drop-out) is a sequence of module enrolment sets, with each set denoting an academic year. Modules can appear along more than one such consecutive set to account for failure and re-attendance.

Of course, a simple enumeration of individual tuition paths does not offer any insight into how populations evolve. Our approach is to simulate the individual legs of each path, allowing for a probabilistic decision at each point in time on what action to take next.

## 2.1. *Drawing the State Space Graph*

A state is the set of modules that a student has selected at an academic year (*all* selected modules up to that point). A transition between states is the selection of modules for the current academic year. That way, we can model the path that a student follows while enrolled for the particular program, where each state represents an academic year and state transitions represent the module registration actions that occur at the start of each academic year.

According to these conventions, the state space for the program we are examining is shown in Fig. 1 (transition probabilities are not shown yet, to avoid cluttering). Numbers indicate module codes (50, 51, 60, 61 and 62). An italicized number annotating a transition line conveys the information that the transition happens upon selection of the particular module (or, modules). As an example (following the dashed lines), note that a student may have registered for module 50 for the first academic year, then moved on to select modules 51 and 60 and finally registered for module 62 for the third year, thus taking three years to move from a start to a sink state (both shown in bold). We consider *state:50,51,60,62* (and its two siblings) as sink states because the master's thesis that follows it must be carried out independent of any module attendance obligations.

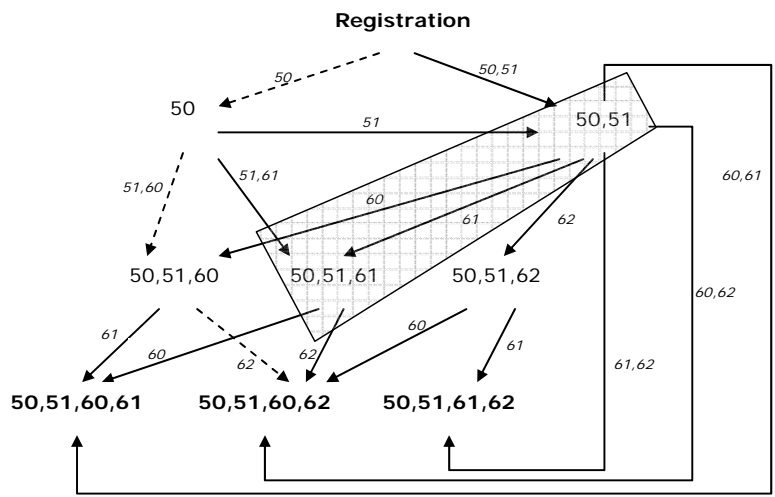

Fig. 1. The state space of module enrolments.

Note that the model in Fig. 1 must be enhanced with transitions from every state to itself (to account for failing a module or a combination of modules and having to repeat it), as well as with transitions from every state to a sink state (to account for dropping out altogether). We do not show such extra transitions to avoid cluttering.

## 2.2. *Calculating the State Transition Probabilities*

The specification of the transition probabilities in our prototype was based on the statistics of the first two years of the program's running. For the senior year modules we substituted default values reflecting that senior year students are very unlikely to fail. Fig. 2 shows a snapshot of the transition probabilities for the state space of our model.

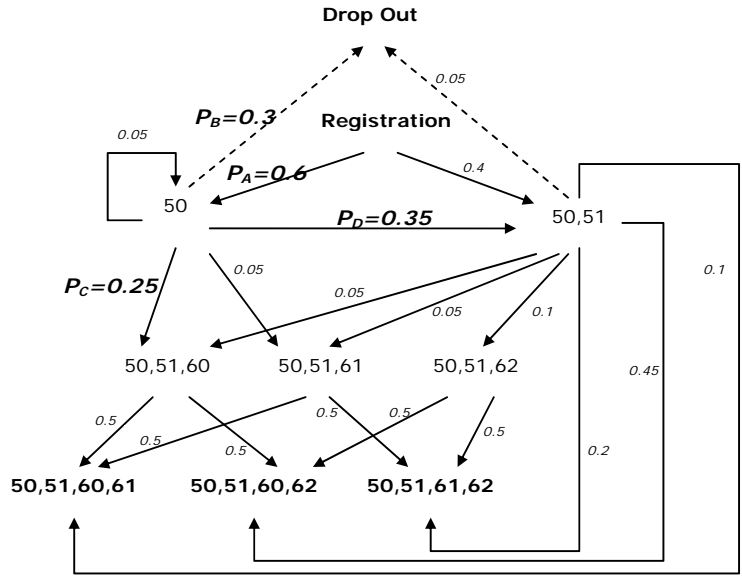

Fig. 2. A snapshot of the transition probabilities of module enrolments.

Probabilities $P_A$, $P_B$, $P_C$ and $P_D$ (shown in larger font) initially seemed to mostly affect the simulation; they corresponded to the most easily observable student paths:
- At the outset of the program, a student may start with enrolling either in module 50 or in modules 50 and 51. The probability to only select module 50 is $P_A$,
- There is a relatively high probability to fail module 50 and to drop out altogether subsequently, $P_B$.
- A successful completion of module 50 is followed by an obligatory enrolment in module 51 and an optional enrolment in a second year unit, usually module 60.
  - The probability to select those two modules is $P_C$.
  - The probability to select just module 51 is $P_D$.

Zero transition and other default probabilities are not shown but there are plenty of them (for example, drop-out at senior modules).

### 2.3. *Using a (Visual) Grammar to Generate the State Space Graph*

For the postgraduate program we are studying, the eventual state space consists of several dozens of states and transitions. As the full representation for a program must account for a range of possibilities on selecting more than a module per year, on deciding the order of module selection, or on deciding which optional modules to select, representation size may scale into hundreds of elements for a program of several modules (as is typical for undergraduate programs in HOU).

At such a model size, the complete specification may be simply unmanageable to draw and the process is prone to errors. A reasonable extension is a modeling notation that captures the precedence constraints between modules and can automatically generate transitions. Instead of introducing that notation formally, we use Fig. 3 as an example that shows the visual graph model for our program.

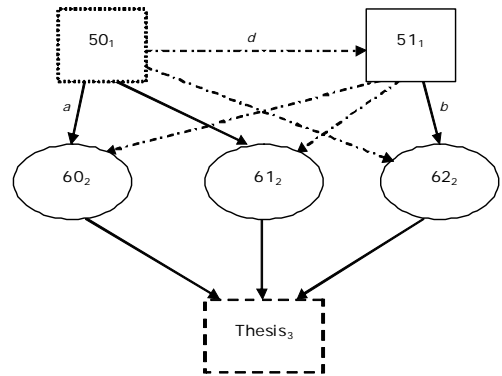

Fig. 3. A high-level visual model of precedence constraints.

Therein, numbers indicate module codes and subscripts indicate the nominal year of the program where a module has been allocated. Edge indices denote precedence constraints (see Table 1 for reference but also note that only a couple of them are shown to avoid cluttering). A solid edge indicates a hard precedence, in the sense that a module cannot be selected (for enrolment) unless its precedent has been completed. A dashed edge indicates a soft precedence, in the sense that a module cannot be selected unless its precedent has at least started. A rectangle indicates a compulsory module; an oval indicates an optional one. A dotted module must be among the first where the students enroll (as is the case with foundation courses); a dashed module must be among the last where the students enroll (as is the case with theses). When an optional module is shown as a precedent to another module, the semantics is that in the final selection of modules, the antecedent may not appear before the precedent (but the precedent may be missing and so will be the antecedent).

Table 1. A snapshot of the list of constraints (precedence constraints are indexed).

| # constraint | Description |
|---|---|
| a | 50 must be successfully cleared before enrolling in either of modules 60 and 61 |
| b | 51 must be successfully cleared before enrolling in module 62 |
| c | Enrolment in any senior module cannot be prior to enrolment in any junior module |
| d | 50 must be the first to be selected |
| e | A thesis is only available as the last module |

We used Prolog to implement a topological ordering that generates all admissible tuition paths on input of a specification as shown in Fig. 3. A few other constraints (not shown above to avoid complicating the notation) were also directly implemented in Prolog (such as allowing students to enroll in at most two modules per year and requiring four modules to advance to a thesis, both of which are not captured in Fig. 3).

**2.4.** *Implementing the Simulation*

We first used the *Extend* simulation software to code the discrete simulation experiment. Extend belongs to the visual programming genre where the programmer selects from a library of modules and creates flows between them; these flows then serve as pathways for discrete simulation objects (simulated students in our case) to move around into the simulated system. While conducting our first experiments with Extend it became glaringly obvious that the size of an Extend implementation and the repetitive nature of the Extend code are both conducive to allowing implementation errors to creep in.

We eventually opted to use Markov chains to implement the calculations that estimate the student populations based on the transition probabilities, instead of simulating *each* student separately. The key tool from the Markov chain toolbox is the formula $v^n = v^{n-1}P = v^1 P^n$ that allows us to express the population at some states (*v* is a state vector) as a function of the transition probabilities matrix (*P*) over a period of defined duration (*n* years, with each year counting as one step).[3]

In our implementation[4] we generated the transition probability matrix automatically from the specification and developed a utility to allow the user to fill-in the probabilities. The implementation also deals effectively with providing an integrated environment within which one can deal with these simulations. Specifically, a Visual Basic application (which can be viewed as a wrapper) was developed to capture the precedence constraints, as shown in Fig. 3. These constraints were then used to generate the corresponding Prolog code that generates the alternative paths. The wrapper was then used to execute the Prolog code and to analyze its output, based on which it created the Markov chain representation of the simulation to follow. The simulation then proceeds by inputting initial student data and by manipulating the transition probabilities. The output is calculated in two distinct way, one by conventional programming and the other by embedding the calculations into MS Excel, so that individual steps can be analyzed and followed-up by seasoned analysts who may want to look for details (MS Excel embedded calculations are quite more time-consuming, however).

**3. On the Validity of the Prototyping Approach**

The estimates obtained from our prototype cannot yet be used anywhere near as a basis for decision making at HOU, though, due to the small sample. Since the fastest a student can expect to proceed to a thesis is two years and we are now just in the fourth year of the program, we believe that the academic year 2010-11 will be the first one where a steady-state as far as student enrolments and tuition paths can be expected, so as to define a starting point for calculating up-to-date and credible transition probabilities.

However, before we get there and as we aspire to eventually build a system for organization wide adoption, we have identified some major issues for our agenda.

First, we might reframe the requirements specification in another notation, most notably that of belief networks.[5] We need to investigate other formalisms that may facilitate the derivation of transition probabilities (merging individual records has been also used in studying alternative counseling courses for social welfare practicing[6]).

We note that our approach is more related to conventional AI planning as opposed to the Operations Research approach, since we do not associate rewards with any intermediate actions and since we treat students equally in terms of goals, regardless of how long they take to graduate.[7] Factoring in rewards would increase complexity and raise the issue of what would be a legitimate reward. Even a subtle reference to rewards makes it necessary to treat optimization aspects of the plan.[8]

Still, our problem does not lie at the core of planning in the AI sense, since we are not interested (yet) in computing optimal plans to a goal.[9] Any sense of optimality, we believe, would take us back into the realm of the conventional OR approach and into Markov decision processes, since we would need to somehow associate each studying route with some measure of quality. Candidate measures could be either person-oriented (for example, a multi-objective criterion of minimizing the number of study years and the expenses due to travel), or system-oriented (for example, using the DEA[10] approach to calculate some measure of system efficiency; incidentally we note that there is no inherent limitation that would not allow our proposed approach to be taken up by conventional universities).

The second issue has to do with the credibility of the transition probabilities. Each academic year sports a different configuration of the student population, with new students arriving each year. A certain organizational memory gradually develops based on students' perceptions about which modules are best to select given one's time available for studying or which modules are easier to follow based on one's earlier enrolments, so associated transition probabilities inevitably change. One may use the statistics of all previous years (implementing a time window), possibly discounting for distant years. Another reasonable option is to just use the statistics for the previous year. Whichever decision one makes, the question that looms is whether this is a decision that must be made at the program or at the university level or, even more importantly, whether this is a decision that statistically speaking matters. This issue should probably be independently studied using a theoretical or applied statistics toolkit and simulation.

The third issue has to do with the structural stability of any given degree program and is the one with by far the most challenging consequences. As years go by, academic program change. When new modules are introduced or some modules are no longer offered, transition probabilities are either calculated afresh or are trimmed. When, however, academic committees decide to move some modules up or down the academic requirements ladder (thus affecting the precedence constraints), then deciding how one might make use of previously calculated transition probabilities for the new configuration seems to be a task that is not well defined. We have not yet investigated the options for dealing with this problem. At present, we believe that the best course would be to use theoretical analysis and statistical simulation to offer us some insight as regards the range of such changes, up to which we might be able to use default values without having to worry about the inevitable errors that incur due to the (educational) system's momentum.

We note that the resolution of the latter two issues may transcend the prototype nature of our current implementation and may require the adoption of specific model description languages or process algebras that facilitate reasoning about simulation and about model consistencies,[11] or the adoption of models that explicitly allow for fuzziness in the specification of probabilities.[12]

Incidentally, the latter two issues also raise the question of the relative importance of the individual model parameters towards the credibility of the final results. This raises the possibility that individual technical steps might prove to be important in an unusual manner. It is quite likely that their most important contribution may be to raise the organizational awareness that simulation is not just number crunching but that, instead, significant analysis and planning is involved when we use constraints to capture a system description.

## 4. Conclusions and future directions

Modeling the size of a student population and how such a population is spread into modules of a program with arbitrary precedence requirements among its modules can surely aid a program's management to plan proactively. In this paper we have shown the key technical ingredients of a system that can provide estimates and we have also presented which aspects of such a system need further research, either in terms of system integration or in terms of robust modeling in circumstances of change.

We acknowledge that population modeling and estimation may be at a wide tangent to policies as practiced by today's universities. We also acknowledge that (even) the strategy consensus required for fielding such systems to actually support university administration may take indefinitely longer than the resolution of the technical and scientific issues that we have already identified. Actually, such consensus is probably of a political rather than a technological nature.[13]

Acknowledging that user acceptance is a key success ingredient of such decision support systems, we have also identified as a future goal to attempt to replicate our experiments for programs that have a much larger history at HOU and see how our methodology measures up with their actual enrolment figures. Referring to earlier approaches,[1] we note that, currently at HOU, we attempt to simulate for medium-term planning in order to gain institutional acceptance and appreciation of the potential for policy formation planning. We need to keep in mind that applied problems which aspire to transcend the nature of basic scientific investigation towards fielding also need some real-world-injected simplification, to manage complexity and to alleviate misunderstandings with users.

None of the techniques and tools we have used aspires to further the state of the art in each of the respective fields. Still, putting these techniques together under a unified architecture and being able to develop models that decision-making users can understand gives rise to the emergence of new modeling tools and methodologies, like the one we have proposed in this paper. This will equip educational managers with the insight they direly need in order to contemplate policy alternatives and to reflect upon past decision based on actual data, which takes the pressure away from using such tools as part of everyday fine-tuning administration.[14] So, even if results never reach the level of statistical significance, it will always be up to innovative managers to use simulation as a tool for exploring what-if scenarios to generate reflections about future policy directions.

**Acknowledgments**

Code and data are available for academic purposes, upon request.